# Exponentially Weighted $\ell_2$ Regularization Strategy in Constructing Reinforced Second-order Fuzzy Rule-based Model

Congcong Zhang[1], Sung-Kwun Oh[2,3] Witold Pedrycz[4,5,6], Zunwei Fu[2], and Shanzhen Lu[7]

[1.] Department of Computer, The University of Suwon, San 2-2 Wau-ri, Bongdam-eup, Hwaseong-si, Gyeonggi-do, 445-743, South Korea
[2.] Research Center for Big Data and Artificial Intelligence, Linyi University, Linyi, 276005, China
[3.] School of Electrical & Electronic Engineering, The University of Suwon, Hwaseong-si, Gyeonggi-do, 18323, South Korea
[4] Department of Electrical & Computer Engineering, University of Alberta, Edmonton T6R 2V4 AB, Canada
[5.] Department of Electrical and Computer Engineering, Faculty of Engineering, King Abdulaziz University, Jeddah, 21589, Saudi Arabia
[6.] Systems Research Institute, Polish Academy of Sciences, Warsaw, Poland
[7.] School of Mathematical Sciences, Beijing Normal University, Beijing, 100875, China

**Abstract**: In the conventional Takagi-Sugeno-Kang (TSK)-type fuzzy models, constant or linear functions are usually utilized as the consequent parts of the fuzzy rules, but they cannot effectively describe the behavior within local regions defined by the antecedent parts. In this article, a theoretical and practical design methodology is developed to address this problem. First, the information granulation (Fuzzy C-Means) method is applied to capture the structure in the data and split the input space into subspaces, as well as form the antecedent parts. Second, the quadratic polynomials (QPs) are employed as the consequent parts. Compared with constant and linear functions, QPs can describe the input-output behavior within the local regions (subspaces) by refining the relationship between input and output variables. However, although QP can improve the approximation ability of the model, it could lead to the deterioration of the prediction ability of the model (e.g., overfitting). To handle this issue, we introduce an exponential weight approach inspired by the weight function theory encountered in harmonic analysis. More specifically, we adopt the exponential functions as the targeted penalty terms, which are equipped with $\ell_2$ regularization ($\ell_2$) (i.e., exponential weighted $\ell_2$, $ew\ell_2$) to match the proposed reinforced second-order fuzzy rule-based model (RSFRM) properly. The advantage of $e\ell_2$ compared to ordinary $\ell_2$ lies in separately identifying and penalizing different types of polynomial terms in the coefficient estimation, and its results not only alleviate the overfitting and prevent the deterioration of generalization ability but also effectively release the prediction potential of the model. The effectiveness of RSFRM is evaluated through 23 machine learning datasets. A comparative analysis shows that the proposed RSFRM results in better performance as well as sound interpretability when compared to the state-of-art methods.

**Introduction**

Fuzzy rule-based model (also known as fuzzy system or fuzzy model) is an advanced modeling architecture based on fuzzy logic, if-then rules, and fuzzy reasoning [1]. A fuzzy rule-based model exploits rules as a means of knowledge representation to formalize the knowledge existed in the model. Moreover, due to its modular architecture, well-developed design methodologies and practices, as well as its advantages in interpretability, it has been used in a wide spectrum of realms such as fuzzy control, pattern analysis, fuzzy decision, time series prediction, robotics, etc. [2-6].

There are two well-known fuzzy rule-based models in dealing with regression issues. The first is the Mamdani fuzzy model, in which the condition and the conclusion parts are represented by fuzzy sets [7]. The second type is the Takagi-Sugeno-Kang (TSK) fuzzy model [8], where the premise is represented by a fuzzy set, and the conclusion is a function. In the case of using the standard pre-defuzzified operation, the Mamdani model can be regarded as a TSK model with constant conclusions. There are lots of different variants of fuzzy rule-based models. In early 2001, Oh and Pedrycz proposed a hybrid neuro-fuzzy model of TSK-type model. The model first uses the clustering method to determine the initial values of the apices of the membership functions employed in the model, and then optimizes and adjusts the positions of the apices of the membership functions through evolutionary algorithm and improved complex algorithm [9]. Besides, the TSK model is also commonly used in classification problems. The TSK model is functionally equivalent to the corresponding radial basis neural network (RBFNN) under certain minor constraints [10]. A concise clustering-based fuzzy model is constructed by combining clustering method and RBFNN, where clustering is used as a learning approach to determine the arguments of the receptive field [11-13]. The output of the receptive field is directly used as the value of the matching degree (firing strength) of the fuzzy model. The final output of the fuzzy model needs to be the weighted average sum of the output of each rule in the model, in which the weighted average can guarantee the semantic integrity of the partition function (the sum of the matching degree is one). These hybrid approaches focus on identifying the parameters of the RBFNN (such as the width and center of the Gaussian Function) without any structural modification. After that, a modified method of replacing the nodes in the activation layer with clusters has been proposed, which directly treats the partition grade from the partition function as the output of the activation layer [14,15,22].

In [16], a novel design method of fuzzy model based on clustering technique is proposed, which enhances the performance of the model by identifying and refining the rules with maximum errors. Kim et al. redesigned the conventional neural network structure based on fuzzy clustering and constructed a context layer by using the output space [17]. The experimental results show that the fuzzy model with context layer can improve the quality of the model. In [18], a TSK-type fuzzy classifier constituted of multiple zero-order (weights between the hidden layer and the output layer are constant) single hidden layer feedforward neural networks (SLFNs) is proposed, which aims to decompose complex problems into several simple problems, reduce memory requirements and computational overhead, as well as maintain classification performance.

Generally, in the conventional TSK-type fuzzy models, constant or linear functions are served as the consequent parts of fuzzy rules (i.e., zero-order TSK type or first-order TSK type models). Also, in some extended TSK-type fuzzy models, the conclusion part is also simplified as much as possible (e.g., the zero-order SLFN is as the consequent part of the rule) [18]. However, some high-order polynomials (e.g., quadratic polynomials) are rarely considered. The reason is that although high-order polynomials can improve the approximation ability of the model, they can also easily lead to the deterioration of the prediction ability of the model (e.g., overfitting). To cope with this issue and boost the generalization ability of the model, we propose a reinforced second-order fuzzy rule-based model, which is implemented with the help of fuzzy clustering (viz., FCM) partition and quadratic polynomial (QP) as well as exponential weighted $\ell_2$ regularization. To be precise, we are looking for the suitable weight function in harmonic analysis theory. In harmonic analysis, there are three weight function classes corresponding to Hardy operator [19], Hardy-Littlewood maximal (HLM) operator [20] and one-sided HLM operator [21] in weighted $\ell_2$-norm, which is the continuous version of weighted $\ell_2$-norm. The rule is that the larger operator matches the smaller weight function class. We will adopt the exponential weights, which matches one-sided HLM, as the target penalty terms. The essential reason is that exponential function (such as $10^x$) is rapidly increasing, which can control the high-order terms in polynomials effectively.

The essential features and novelty of our work can be enumerated as follows. First, a novel reinforced second-order fuzzy rule-based model is proposed based on FCM partition and quadratic polynomial (QP) as well as exponential weighted $\ell_2$ regularization ($ew$-$\ell_2$) to deal with regression problems. Second, inspired by the weight function theory in harmonic analysis, an exponential weighted $\ell_2$ regularization is designed to mitigate overfitting elaborately. Different from ordinary $\ell_2$, $ew$-$\ell_2$ can identify and penalize the terms of different types of polynomials in coefficient estimation. Third, QP and $ew$-$\ell_2$ are used in collaboration, and its result can effectively improve the prediction accuracy and stability of the model.

In the sequel, the main contribution of this study is that the effective strategy through exponential weighted $\ell_2$ regularization ($ew$-$\ell_2$) combined with FCM partition and QP leads to better prediction accuracy as well as sound interpretability in constructing fuzzy rule-based model based on a series of experimental comparative studies.

The paper is arranged in the following sections. First, Section 2 presents the architecture of the reinforced second-order fuzzy rule-based model. The learning mechanism with exponential weighted $\ell_2$ regularization is discussed in Section 3. The design steps of RSFRM are reported in Section 4. The effectiveness of RSFRM is verified in Section 5. Section 6 concludes and outlines future work.

## 2. Architecture of reinforced second-order fuzzy rule-based model
### 2.1. Takagi-Sugeno-Kang fuzzy rule-based model

Takagi–Sugeno-Kang (TSK) fuzzy model (also known as TS or Sugeno fuzzy model) is a classic fuzzy rule-based model (refer to Fig. 1(a)), which aim to devise a systematic fashion to generating fuzzy rules from a finite set of input-out data pairs [8,10]. A representative form of fuzzy rule in TSK fuzzy model is:

$$R_i: IF\ \underbrace{x_1\ is\ L_{i1}\ and\ \ldots\ x_n\ is\ L_{in}}_{\text{antecedent part}}\ THEN\ \underbrace{g_i = f_i(\mathbf{x}_k)}_{\text{consequent part}} \#(1)$$

where $L_{i1}$, $L_{i2}$,…, $L_{in}$ represent the linguistic fuzzy terms in the antecedent part, and $f_i(\mathbf{x}_k)$ denotes the crisp function in the consequent part, $\mathbf{x}_k = [x_1, x_2,\ldots, x_n]^T \in \Re^n$ stands for the $k^{th}$ input vector. Generally, $f_i(\mathbf{x}_k)$ is the zero-order or first-order polynomial of the input variables. TSK fuzzy model can be expressed as an equivalent feedforward neural network, namely the fuzzy neural network (FNN), as displayed in Fig. 1(b) [10, 17, 18]. In Fig. 1(b), the $1^{st}$ layer stands for the input layer, "**L**" node in the $2^{nd}$ layer denotes the linguistic fuzzy term denoted by membership function, the output of "**M**" node in the $3^{rd}$ layer represents matching degree $h_i(\mathbf{x}_k)$ and which can be calculate as:

$$h_i(\mathbf{x}_k) = T\ (L_{i1}(x_1),L_{i2}(x_2),L_{i3}(x_3),\ldots,L_{in}(x_n)) \#(2)$$

where $T$ denotes the $T$-norm operator (e.g., minimum and product are commonly used here). $L_{in}$ indicates the linguistic fuzzy term of the $n^{th}$ input variable to the $i^{th}$ rule. The "**N**" node in the $4^{th}$ layer represents the normalization operation. The node of layer 5 yield local output (model) by multiplying the consequent part and their associated antecedent part.

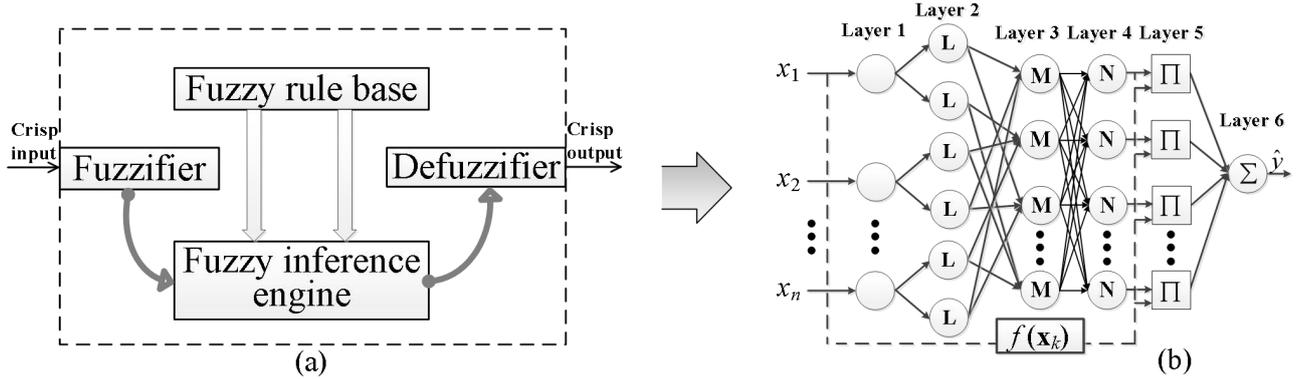

**Fig. 1.** Architecture of TSK fuzzy model. (a) Standard fuzzy rule-based model. (b) Equivalent FNN architecture of TSK fuzzy model.

The overall output of FNN is a combination of all the local models:

$$\hat{g}_k(\mathbf{x}_k) = \sum_{i=1}^{c} f_i(\mathbf{x}_k)\left(h_i(\mathbf{x}_k)/\sum_{r=1}^{c} h_r(\mathbf{x}_k)\right) = \sum_{i=1}^{c} f_i(\mathbf{x}_k)\tilde{h}_i(\mathbf{x}_k)\ \#(3)$$

where $\hat{g}_k$ denotes the $k^{\text{th}}$ output of the model, $c$ represents the number of fuzzy rules, $\tilde{h}_i(\mathbf{x}_k)$ indicates the normalized matching degree of the $k^{\text{th}}$ data to the $i^{\text{th}}$ rule.

### 2.2. Architecture of the proposed second-order fuzzy rule-based model

The essence of the fuzzy rule-based model is embodied in the idea of "divide and conquer". That is, the antecedent parts of the fuzzy rule define the local regions (models), and the consequent parts describe the behavior of the region through different components. Therefore, when exploiting the model to some local regions, these regions need to be guardedly delimited. An information granulation (Fuzzy C-Means, FCM) method is utilized to analyze and reveal data distribution on the input space and form local regions in this study.

The local regions of the input space are defined by some linguistic fuzzy terms (fuzzy sets) generated by FCM clustering algorithms. Given a finite set input variables $\mathbf{X} = [\mathbf{x}_1, \mathbf{x}_2, \mathbf{x}_3, \ldots, \mathbf{x}_N] \in \Re^{n \times N}$, $\mathbf{x}_k \in \Re^n$, the loss function of FCM is as follows:

$$J_m = \sum_{i=1}^{c}\sum_{k=1}^{N}(u_{ik})^m\ ||\mathbf{x}_k - \mathbf{v}_i||^2,\ 1 < m < \infty \#(4)$$

where $c$ indicates the number of clusters, as well as the number of rules of the proposed model. $\mathbf{v}_i$

stands for the $i^{th}$ prototype. $||\mathbf{x}_k - \mathbf{v}_i||$ represents the Euclidean Metric between the $k^{th}$ data and the $i^{th}$ prototype. $m$ denotes the fuzzification coefficient is used to control the specificity of the underlying linguistic information (the typical value of $m$ is 2). $u_{ik}$ represents the membership grade of the $k^{th}$ pattern in the $i^{th}$ cluster (fuzzy set).

Moreover, $u_{ik}$ needs to satisfy the following restrictions:

$$Res = \left\{ 0 \leq u_{ik} \leq 1, \sum_{i=1}^{c} u_{ik} = 1 \; \forall k, 0 < \sum_{k=1}^{N} u_{ik} < N \; \forall i \right\} \#(5)$$

By minimizing the loss function, we obtain $u_{ik}$ and $\mathbf{v}_i$:

$$u_{ik} = \frac{1}{\sum_{j=1}^{c} \left( \frac{||\mathbf{x}_k - \mathbf{v}_i||}{||\mathbf{x}_k - \mathbf{v}_j||} \right)^{\frac{2}{(m-1)}}} \#(6)$$

$$\mathbf{v}_i = \frac{\sum_{k=1}^{N} u_{ik}^m \mathbf{x}_k}{\sum_{k=1}^{N} u_{ik}^m} \#(7)$$

The values of $u_{ik}$ and $\mathbf{v}_i$ are updated by the FCM algorithm through iteration (6) and (7).

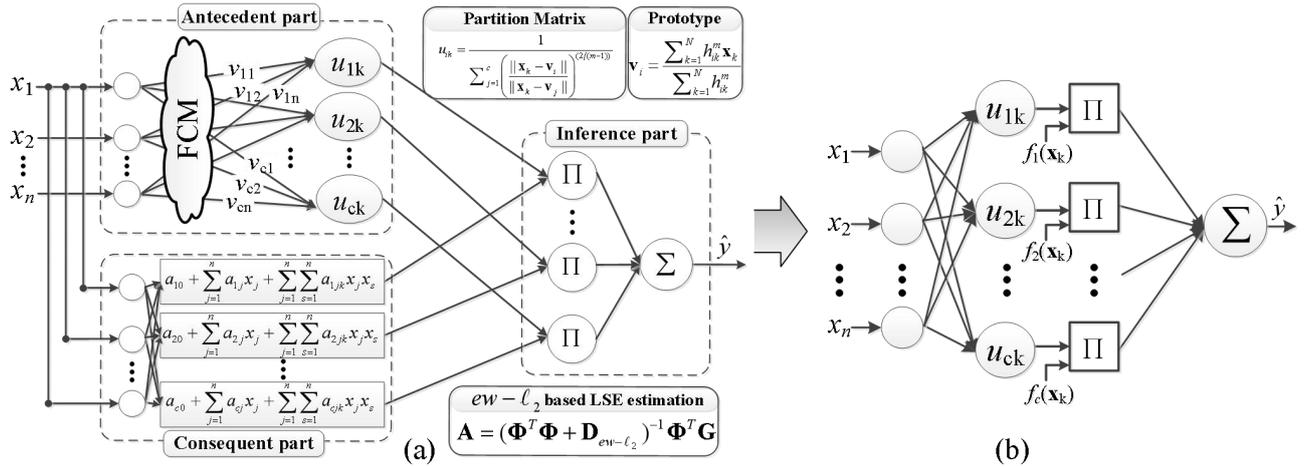

**Fig. 2.** Architecture of the reinforced second-order fuzzy rule-based model (RSFRM). (a) Modular architecture of RSFRM. (b) Equivalent FNN architecture of RSFRM.

The proposed model implements a modular fuzzy architecture through three parts (antecedent, consequent, and inference), as displayed in Fig. 2(a). The antecedent and consequent parts are related to the constitution of fuzzy rules and their subsequent analysis. The inference part involves the aggregation of multiple fuzzy rules. Unlike the conventional TSK-type fuzzy model (refer to Fig. 1(b)), the structure of the proposed model is more compact and concise. The second and third layers in Fig. 1(b) are compressed into one layer (refer to Fig. 2(b)) (that is, the output of the partition grade is directly used as the matching degree). In addition, since the partition grade needs to satisfy the constraint conditions (Eq. (5)), the fourth layer in Fig. 1 (b) is omitted. Fig. 2 illustrates the modular architecture and the Equivalent FNN architecture of the reinforced second-order fuzzy rule-based model (RSFRM).

In the conventional TSK-type fuzzy model, zero-order or first-order polynomials (viz., constant or linear function) are generally selected as the consequents of fuzzy rules, and high-order polynomials (e.g., quadratic polynomials) are rarely considered. In fact, quadratic polynomials ($f_i(\mathbf{x}_k)$) have a sound description and interpretation of the local regions (models) divided by the antecedents, but they are prone to overfitting.

$$f_i(\mathbf{x}_k) = a_{i0} + \sum_{j=1}^{n} a_{ij}x_j + \sum_{j=1}^{n}\sum_{s=1}^{n} a_{ijk}x_j x_s, (\mathbf{x}_k \in \Re^n, 1 < i < c) \#(8)$$

where $a_{i0}$, $a_{ij}$, $a_{ijk}$ represent the coefficients of polynomial, $n$ denote the amount of input variables.

In order to cope with this problem and enhance the prediction accuracy of RSFRM, exponential weighted $\ell_2$ regularization and quadratic polynomial are used in collaboration. The output of RSFRM can be rewritten as follows:

$$\hat{g}_k(\mathbf{x}_k) = \sum_{i=1}^{c} f_i(\mathbf{x}_k) u_{ik} \#(9)$$

where $\hat{g}_k$ stands for the $k^{\text{th}}$ output of the model, $c$ stands for the number of fuzzy rules, $u_{ik}$ represents the partition grade.

## 3. Learning mechanism with exponential weighted $\ell_2$ regularization
### 3.1. Least square error estimation for the proposed model

The proposed RSFRM uses the Least Square Error (LSE) estimation to calculate the coefficients of the consequent part (polynomial) of the fuzzy rule. The objective function of LSE is to minimize the sum of the squared error of the observed (original) output and the model output, as in Eq. (10):

$$J = \sum_{k=1}^{N} (g_k - \hat{g}_k)^2 = \sum_{k=1}^{N} (g_k - \sum_{i=1}^{c} f_i(\mathbf{x}_k) u_{ki})^2 \#(10)$$

where $N$ denotes the number of data patterns, $g_k$ and $\hat{g}_k$ represent the $k^{th}$ real and model output, respectively. $f_i(\mathbf{x}_k)$ denotes the consequent part of the $i^{th}$ rule, $u_{ki}$ indicates the partition grade of the $k^{th}$ pattern to the $i^{th}$ prototype.

By taking the derivative of objective function (10), the general expression of the coefficient vector can be obtained:

$$\mathbf{A} = (\Phi^T \Phi)^{-1} \Phi^T \mathbf{G} \#(11)$$

where the specific forms of $\mathbf{A} \in \Re^{1 \times (c(n^2+3n+2)/2)}$, $\Phi \in \Re^{N \times (c(n^2+3n+2)/2)}$, and $\mathbf{G} \in \Re^{1 \times n}$ are as follows (illustrated by the example of quadratic polynomial):

$$\mathbf{G} = \begin{bmatrix} u_{11} & \cdots & u_{c1} & x_{11}u_{11} & \cdots & x_{n1}u_{c1} & x_{11}^2 u_{11} & \cdots & x_{n1}^2 u_{c1} \\ u_{12} & \cdots & u_{c2} & x_{12}u_{12} & \cdots & x_{n2}u_{c2} & x_{12}^2 u_{12} & \cdots & x_{n2}^2 u_{c2} \\ \vdots & \vdots & \vdots & \vdots & \vdots & \vdots & \vdots & & \vdots \\ u_{1N} & \cdots & u_{cN} & x_{1N}u_{1N} & \cdots & x_{nN}u_{cN} & x_{1N}^2 u_{1N} & \cdots & x_{nN}^2 u_{cN} \end{bmatrix} \#(12)$$

$$\mathbf{A} = [a_{1,0} \cdots a_{c,0}, a_{1,1} \cdots a_{c,n}, a_{1,(n+1)} \cdots a_{c,(n^2+3n)/2}]^T, \mathbf{G} = [g_1, g_2, \cdots, g_N]^T \#(13)$$

### 3.2. Exponential weighted $\ell_2$ regularization

Theoretically, as the complexity of the model increases, the ability of the model to approximate the data becomes better, and the predictive performance of the model increases. However, the prediction performance of the model actually decreases as the model complexity increases excessively (i.e., overfitting). For the proposed fuzzy model, its complexity can be considered from two aspects. (a) The number of fuzzy rules. The fuzzy rules are mainly determined by the partition of the input space, which is a pivotal indicator to assess the complexity of the neuro-fuzzy model [23-24]. Also, the number of fuzzy rules is hyperparameter and needs to be specified in advance. (b) The type of consequent part of the fuzzy rules. The consequent part describes the behavior of the local region through different components (e.g., crisp functions) [10]. Although the more complex components can describe the local region in detail, the risk of overfitting faced by the model is higher. The consequent part of the rules of the proposed model is constructed by polynomial functions. Under the premise that the order and variables of the polynomial are determined, only its coefficients can affect the performance and stability of the model. The simpler the coefficient, the more stable the model becomes. In the simplest case, all coefficients are directly set to zero, and the output of the model is always zero. The model is simple, but it is meaningless because it cannot be effectively predicted.

As mentioned before, in the proposed model, LSE is used to minimize the defined objective function (i.e., Eq. (10)) to compute the coefficients of the consequent parts of the fuzzy rules. In the process of coefficient estimation, multiple factors including overtraining could lead to the estimation of unstable coefficients with large deviations between the coefficients, resulting in a poor generalization ability of the model. Generally, we consider using $\ell_2$ regularization ($\ell_2$) to improve the weakness of LSE [25-27]. $\ell_2$ is a technique for analyzing and alleviating overfitting of multivariate regression. This approach helps reduce the deviation between the coefficients and prevents the generalization ability from deteriorating. By adding the $\ell_2$ regularization parameter (penalty term) to the objective function applied in the current fuzzy model, the $\ell_2$ regularization can be easily applied, as shown below

$$J_{\ell_2} = \sum_{k=1}^{N}(g_k - \hat{g}_k)^2 + \lambda \sum_{p=0}^{(n^2+3n)/2} \sum_{i=1}^{c}(a_{ip})^2 \quad (14)$$

where $N$ denotes the number of data, $g_k$ and $\hat{g}_k$ indicate the $k^{th}$ real and model output, respectively. $\lambda$ stands for the penalty term, $c$ stands for the number of fuzzy rules, $n$ denotes the number of input variables, and $a_p$ are the coefficients of the quadratic polynomial.

We can obtain the extended coefficient vector expression by taking the partial derivative of (14). Compared with the general estimation method, the regularized estimation method adds a positive number on the diagonal of the design matrix ($\Phi^T\Phi$) to make the matrix nonsingular.

$$\mathbf{A}_{\ell_2} = (\Phi^T\Phi + \lambda \mathbf{I})^{-1}\Phi^T\mathbf{G} \quad (15)$$

However, the ordinary $\ell_2$ penalizes all coefficients uniformly [26-27], that is, adding the same penalty factor to each entry of the diagonal of the design matrix. In this way, it may be unfair to treat all entries on the diagonal of the design matrix in the same way. It does not consider the different impacts of different types of polynomial terms on the performance of the fuzzy model.

As shown in Fig. 3, the diagonal of the design matrix contains three types of matrix entries of different orders, such as zero-order entries, quadratic entries, and biquadratic entries, composed of different polynomial terms and their corresponding matching degrees. Obviously, these three types of entries have different effects on the coefficient estimation as well as the performance of the model. In the consequent parts of the fuzzy rules, the higher-order terms of the polynomials can describe the behavior within the local region defined by the antecedent parts of fuzzy rules in more detail, which help the model to exhibit a stronger approximation ability, but exposes the model to a higher risk of overfitting. Therefore, in the process of regularization, it is unreasonable to treat the terms of different types (orders) in polynomial as equivalent, and use the same penalty terms to punish uniformly the coefficients of terms of different types. The unified penalty limits the ability of consequent parts of fuzzy rules composed of polynomial terms of different orders to properly describe (represent) the behavior within local regions. Inspired by the weight function theory of harmonic analysis, we propose exponential weighted $\ell_2$ regularization ($ew\text{-}\ell_2$) to achieve a more precise "penalty" (or "targeted penalty"). More specifically, different penalty terms are selected according to polynomial terms of different types, so that the coefficients of polynomial terms of different types can be adjusted (punished) more fairly and reasonably.

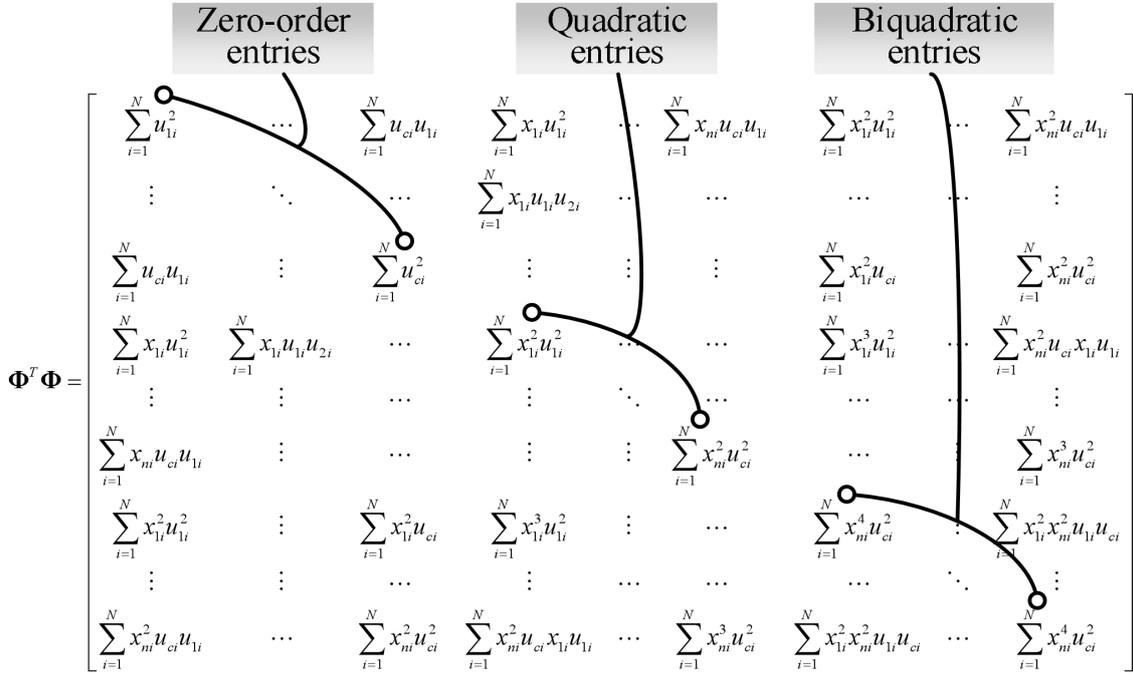

**Fig. 3.** Schematic of the design matrix.

Ordinary $\ell_2$ can be regarded as a critical (special) form of $ew\text{-}\ell_2$, in which all the penalty terms of $ew\text{-}\ell_2$ are equal. Compared with the ordinary $\ell_2$, $ew\text{-}\ell_2$ can more accurately slow down the excessive description of behavior with local regions by polynomials with high-order terms through different penalty terms, which can promote the consequent parts (denoted by polynomials) can more reasonably represent the behavior within local regions during the training phase and effectively predict the unknown behavior within local regions during the test phase. In other words, $ew\text{-}\ell_2$ can provide opportunities to improve generalization ability by establishing a effective trade-off between the training and testing of the model, and reasonable release the representation (description) ability of high-order polynomials within local regions, which not only alleviates the overfitting of the model caused by overtraining and other factors, but also effectively boosts the prediction performance of the model.

The objective function considering $ew\text{-}\ell_2$ can be augmented as:

$$J_{ew-\ell_2} = \sum_{k=1}^{N}(g_k - \hat{g}_k)^2 + \lambda_1 \sum_{p_1=0}^{0}\sum_{i=1}^{c}(a_{ip_1})^2 + \lambda_2 \sum_{p_2=1}^{n}\sum_{i=1}^{c}(a_{ip_2})^2 + \lambda_3 \sum_{p_3=n+1}^{(n^2+3n)/2}\sum_{i=1}^{c}(a_{ip_3})^2,$$
$$\text{subject to } 0 < \lambda_1 < \lambda_2 < \lambda_3 \#(16)$$

Here, $N$, $g_k$, $\hat{g}_k$, $c$, $n$, and $a(a_{ip_1}, a_{ip_2},...)$ have the same meaning as in (14).

Eq. (16) can be represented in the form of matrix, as follows:

$$J_{ew-\ell_2}(\mathbf{A}) = (\mathbf{G} - \Phi\mathbf{A})^T(\mathbf{G} - \Phi\mathbf{A}) + \mathbf{D}_{ew-\ell_2}\mathbf{A}^T\mathbf{A} \#(17)$$

where $\mathbf{D}_{ew-\ell_2} \in \Re^{q \times q}$ is a diagonal matrix containing $\lambda_1$, $\lambda_2$, $\lambda_3$ three types of entries, $q = c(n^2+3n+2)/2$.

We take the derivative of $J_{ew-\ell_2}$:

$$\frac{\partial J_{ew-\ell_2}(\mathbf{A})}{\partial \mathbf{A}} = \nabla_{\mathbf{A}}\big(\mathbf{G}^T\mathbf{G} - \mathbf{G}^T\Phi\mathbf{A} - \mathbf{A}^T\Phi^T\mathbf{G} + \mathbf{A}^T\Phi^T\Phi\mathbf{A} + \mathbf{A}^T\mathbf{D}_{ew-\ell_2}\mathbf{A}\big) \#(18)$$
$$= 2\big(\mathbf{0} - \Phi^T\mathbf{G} + \Phi^T\Phi\mathbf{A} + \mathbf{D}_{ew-\ell_2}\mathbf{A}\big)$$

Let Eq. (16) be equal to $\mathbf{0}$, the analytic solution to the polynomial coefficients is expressed as:

$$\mathbf{A} = (\Phi^T\Phi + \mathbf{D}_{ew-\ell_2})^{-1}\Phi^T\mathbf{G} \#(19)$$

## 4. Design procedure of the reinforced second-order fuzzy rule-based model

Overall, the design procedure of the reinforced second-order fuzzy rule-based model includes the following steps:

**[Step 1] Form training and test datasets.**

Divide the given dataset into two blocks, namely the training and test datasets. The training data is used to establish the model, and the remaining data is utilized to estimate the predictive performance of the model.

**[Step 2] Calculate the matching degree of each fuzzy rule.**

Information granulation (e.g., Fuzzy C-Means) technique is utilized to analyze and reveal the data distribution of the input space to form the local regions. At the same time, the matching degree (which is denoted by partition grade) of each fuzzy rule in the corresponding local region are calculated by using (6) and (7).

**[Step 3] Estimate the coefficients of the consequent part in each fuzzy rule.**

Coefficients of the consequent part in each fuzzy rule are computed by LSE estimation using (19). Here, LSE and $ew\text{-}\ell_2$ are used together. Before using, we need to set the parameters of $ew\text{-}\ell_2$ in advance.

**[Step 4] Compute the predicted output of the proposed RSFRM by using (9).**

## 5. Experimental studies
## 5.1. Experimental setup

We use training error and test error to represent the performance of each model (including control and comparative models) on training data and test data. The performance of the model is mainly quantified as Root Mean Square Error (RMSE) and Mean Square Error (MSE), as in Eq. (20):

$$\text{Training/Test error} = \begin{cases} \sqrt{\frac{1}{N}\sum_{j=1}^{N}(y_j - \hat{y}_j)^2}, & \text{(RMSE)} \\ \frac{1}{N}\sum_{j=1}^{N}(y_j - \hat{y}_j)^2, & \text{(MSE)} \end{cases} \quad \#(20)$$

The basic comparative models of the experiment mainly include the zero-order fuzzy rule-based model (zero-order FRM), first-order FRM, second-order FRM and second-order FRM with $\ell_2$. The order indicates the types of the conclusion parts of the fuzzy rules such as constants, linear functions and quadratic polynomials. All experiments are based on $k$-fold cross-validation, and each dataset is split into $k$ chunks, of which one of chunks is used for testing, and the rest is used for training. This process is repeated $k$ times. $k$ is set to 5 in this section. The experimental parameter setting is summarized in Table 1, which is determined via $k$-fold cross-validation method. The reason for choosing these specific values is that we need to consider the possibility of studying the performance of RSFRM in a variety of scenarios.

**Table 1** Specification of the parameters

| Parameters | | Values |
| --- | --- | --- |
| Fuzzy C-Means parameters | Fuzzification coefficient ($m$) | 2.0 |
| | Number of clusters/ rules ($c$) | 2, 4, 6 ,8, 10 |
| Exponential weighted $\ell_2$ regularization parameters | Penalty terms ($\lambda_1, \lambda_2, \lambda_3$) subject to $0 < \lambda_1 < \lambda_2 < \lambda_3$ | $10^{-8}, 10^{-6}, 10^{-4}, 10^{-2}, 10^{-1}, 10^0, 10^1, 10^2, 10^3$ |

### 5.2. Specific evaluations of the proposed model
#### 5.2.1. Synthetic data

We consider a nonlinear function containing two variables, whose mathematical description is as follows:

$$y(\mathbf{x}) = \Phi(x_1, x_2) = 1.9(1.35 + e^{x_1} e^{x_2} \sin(13(x_2 - 0.6)^2) \sin(7x_1)) \#(21)$$

where the domain of definition of input $(x_1, x_2)$ is [0, 1]. Fig. 4(a) visualizes the set of solutions of (21). Randomly select 500 pairs of input-output samples from the solution set as training and test datasets, where Fig. 4(b) displays the spatial distribution of the selected data.

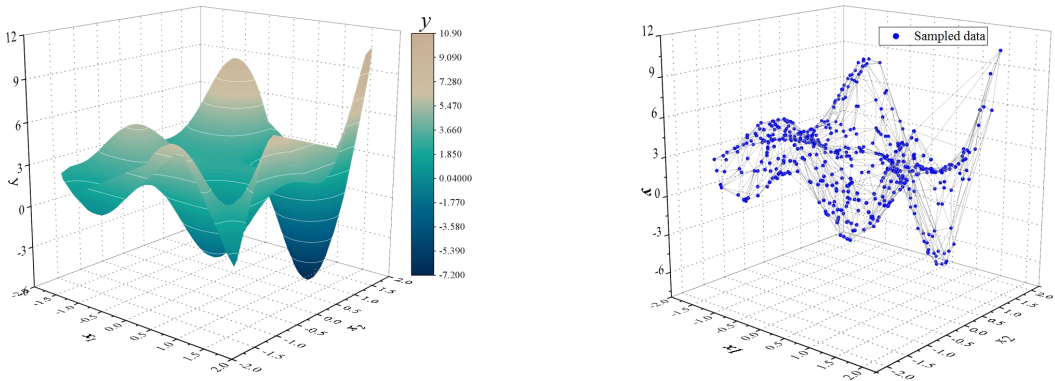

(a) Solution set of Synthetic data.      (b) 500 pairs sampled data for experiment.

**Fig. 4.** Synthetic data.

Table 4 shows the performance (which is denoted by training/ test error) of the control (proposed)

and the comparative models including the zero/ linear/second-order fuzzy rule model (FRM) as well as the second-order FRM with regularization. The performance values are reported as average and its associated standard deviation (STD). The items in bold denote the best performance. We utilize MSE/2 as the performance index. Obviously, compared with zero-order and first-order FRM, second-order FRM has better approximation and prediction capabilities. For the three second-order FRMs with $\ell_2$ (which use different penalty terms), their training error is the same as that of the second-order FRM without $\ell_2$, but their test error is slightly worse. This means that using ordinary $\ell_2$ does not improve the performance of the model. Although the proposed model shares the same training error as the second-order FRM and its variants with $\ell_2$, its generalization ability is improved by using $ew$-$\ell_2$.

**Table 2** Experimental results of comparative and proposed models. (The performance values in this table should be multiplied by $10^{-2}$.)

| Models | No. of rules ($c$) | Penalty terms ($\lambda_1, \lambda_2, \lambda_3$) | Training error Mean | STD | Test error Mean | STD |
|---|---|---|---|---|---|---|
| Zero-order FRM | 6 | N/A | 208.9 | 23.92 | 217.8 | 63.25 |
| Linear-order FRM | 10 | N/A | 22.98 | 1.922 | 32.51 | 13.21 |
| Second-order FRM | 10 | N/A | 2.977 | 0.802 | 5.914 | 3.535 |
| Second-order FRM with $\ell_2(\lambda_1/\lambda_2/\lambda_3)$ | 10 | $10^{-8}$ | 2.977 | 0.802 | 6.097 | 3.662 |
|  | 10 | $10^{-6}$ | 2.977 | 0.802 | 6.097 | 3.662 |
|  | 10 | $10^{-4}$ | 2.977 | 0.802 | 6.097 | 3.656 |
| RSFRM (Proposed model) | 10 | $(10^{-8}, 10^{-6}, 10^{-4})$ | **2.977** | **0.802** | **5.779** | **3.858** |

($\ell_2$: $\ell_2$ regularization, RSFRM: reinforced second-order fuzzy rule-based model)

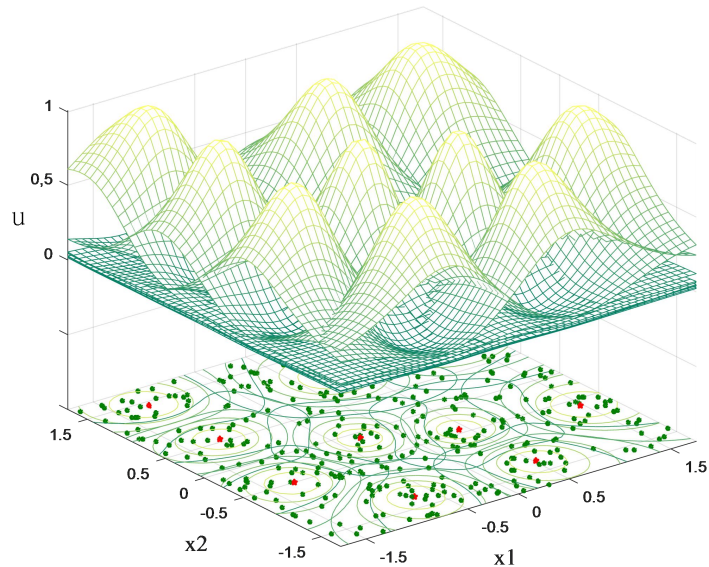

**Fig. 5.** Description of partition of input space and antecedent parts of fuzzy rules formed by FCM.

Fig. 5 visualizes the partition of input space and the antecedent parts of fuzzy rules formed by FCM. The input space is divided by ten local regions (subspaces), and those values are served as matching degrees to estimate coefficients of the consequent parts of the fuzzy rule. Fig. 6 displays the comparison between the target (actual) output and the model output. Fig. 6(a) visualized 500 actual input-output points, while in Fig. 6(b), the blue tetrahedron stands for the model's training output, and the red cube denotes the model's predicted output.

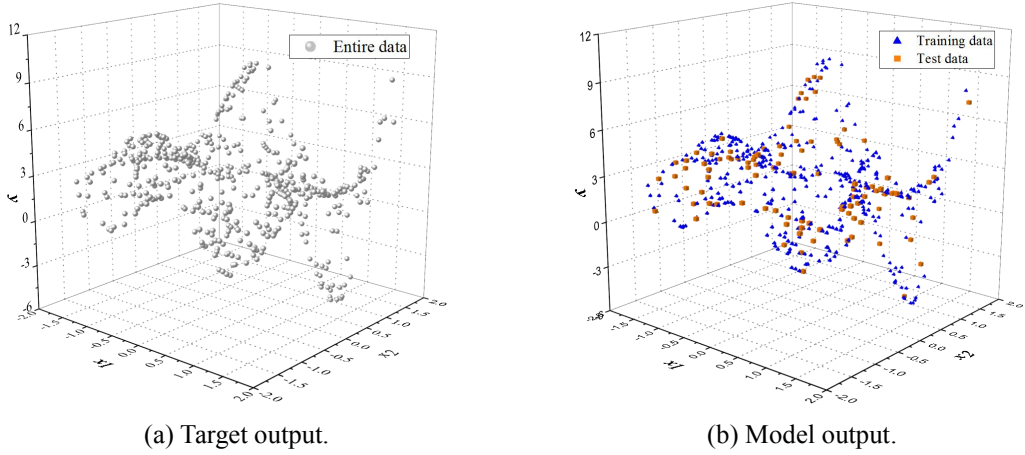

(a) Target output.  (b) Model output.

**Fig. 6.** Comparison of target output data and model output.

Table 3 Experimental results of comparative and proposed models (MPG)

| Models | No. of rules ($c$) | Penalty terms ($\lambda_1, \lambda_2, \lambda_3$) | Training error Mean | STD | Test error Mean | STD |
|---|---|---|---|---|---|---|
| Zero-order FRM | 8 | N/A | 3.908 | 0.192 | 3.958 | 0.381 |
| Linear-order FRM | 2 | N/A | 2.778 | 0.035 | 2.969 | 0.109 |
| Second-order FRM | 8 | N/A | 0.595 | 0.044 | 228.8 | 178.4 |
| Second-order FRM with $\ell_2(\lambda_1/\lambda_2/\lambda_3)$ | 8 | $10^{-1}$ | 1.807 | 0.046 | 2.867 | 0.287 |
|  | 8 | $10^0$ | 2.365 | 0.048 | 2.963 | 0.113 |
|  | 8 | $10^1$ | 4.060 | 0.047 | 4.381 | 0.321 |
| RSFRM (Proposed model) | 8 | ($10^{-1}, 10^0, 10^1$) | **2.370** | **0.047** | **2.702** | **0.178** |

### 5.2.2. Automobile Miles Per Gallon data (MPG)

Automobile MPG data includes 7 input variables and 392 input-output data pattern pairs. The output is the fuel consumption of the automobile represented by miles per gallon. Table 3 reports the performance (viz., training and test errors) of the control and the comparative models. RMSE is used as the performance index. The value of the performance is described in terms of mean and its related standard deviations (STD). Obviously, compared with zero-order FRM and linear-order FRM, second-order FRM has a tiny training error. However, compared with the fitting advantage of second-order FRM in training data, its generalization ability (test error) is severely deteriorated. Although the $\ell_2$ regularization ($\ell_2$) technique can alleviate the overfitting problem of the second-order FRM, and it does not give full play the prediction performance of the quadratic function (QP) well. Compared with ordinary $\ell_2$, $ew\text{-}\ell_2$ can separately adjust (punish) different types of polynomial terms, so it can improve the predictive ability of the model by reasonably release the ability of QP to describe the behavior within local region. The optimal predictive performance of the proposed RSFRM is 2.702±0.178, which is significantly improved than that of other comparative models.

We use the logarithm of the sum of the squared coefficients (LSSC) of the consequent part of fuzzy rules to look into the effect of regularization techniques ($\ell_2$ and $ew\text{-}\ell_2$) on the change of coefficients (the deviation among coefficients).

$$LSSC = \log_e \left( \sum_{p=1}^{(n^2+3n)/2} \sum_{i=1}^{c} (a_{ip})^2 \right) \quad \#(22)$$

Fig. 7 shows the effect of regularization techniques on the coefficients and performance of the model. Without regularization, the training error is close to zero, and thereby the approximation ability reaches the ideal result. But there is a large interval between the generalization ability and approximation ability of this model. When ordinary $\ell_2$ is used, the LSSC of the model decreases with the increase of the penalty term ($\lambda$), and the training error of the model increases with the increase of the penalty term (without $\ell_2$, the penalty term can be regarded as zero). Regularization can strengthen the generalization ability of the model, but there is no definite relationship between the penalty term and test error. Compared with the three models using ordinary $\ell_2$, the LSSC and training errors of the proposed model using $ew\text{-}\ell_2$ are close to the one with the minimum test error, but the generalization ability (test error) of the proposed model is superior to that of this comparative model. This is because $ew\text{-}\ell_2$ penalizes different polynomial terms separately, which can sound balance the relationship between the model's approximation ability and prediction ability through fine shrinking of the deviation among coefficients. In other words, $ew\text{-}\ell_2$ can better boost the prediction performance of the model when compared with ordinary $\ell_2$.

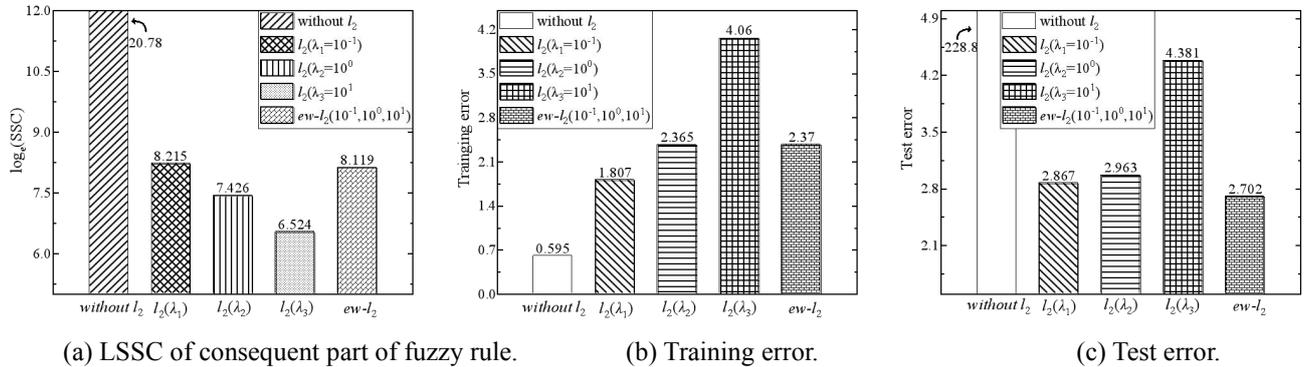

(a) LSSC of consequent part of fuzzy rule.　　(b) Training error.　　(c) Test error.

**Fig. 7.** Effects of regularization ($\ell_2$ and $ew\text{-}\ell_2$) on the coefficients and performance of models (MPG).

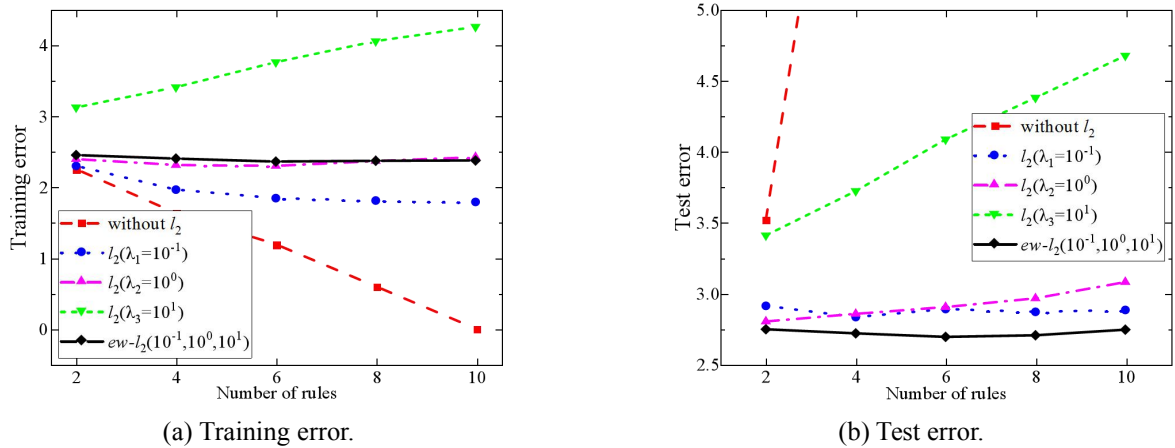

(a) Training error.　　(b) Test error.

**Fig. 8.** Effects of complexity (number of rules) on the performance of models (MPG).

Fig. 8 shows the effect of the complexity (denoted by the number of fuzzy rules) on the performance of the model. In the case of $\ell_2$ is not used, the training error of the model decreases sharply with the increase of the complexity, but its test error deteriorates seriously. When using ordinary $\ell_2$, a suitable λ (e.g., $10^{-1}$) can slow down the overfitting of the model and improve the generalization ability of the model. In contrast, an excessively large λ (e.g., $10^1$) may cause the training error and test error of the model to deteriorate simultaneously. The proposed model with $ew\text{-}\ell_2$ can respectively punish coefficients of polynomial terms of different types, which not only reduces the overfitting of the model but also releases the representation ability of quadratic function within the local region. Compared with other comparative models, the proposed model with $ew\text{-}\ell_2$ exhibits better prediction ability and stability under different complexity situations.

Fig. 9 displays the error comparison between the model output and the target output, where the blue line indicates that the error between the two outputs is zero. Compared with the three comparative models, the output positions of RSFRM are closer to the blue line (baseline). Table 4 provides a comparison with existing models in the literature.

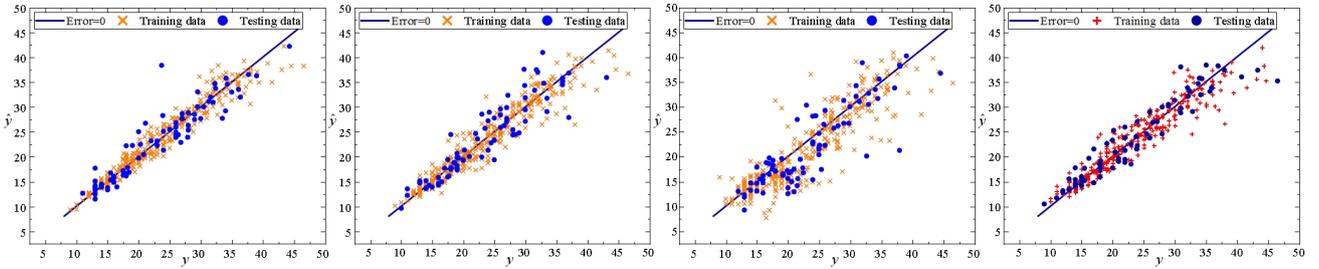

(a) SFRM with $\ell_2(\lambda_1)$  (b) SFRM with $\ell_2(\lambda_2)$  (c) SFRM with $\ell_2(\lambda_3)$  (d) Proposed model

**Fig. 9.** Comparison of target output and model output for MPG (SFRM: second-order fuzzy rule-based model).

**Table 4** Results of comparative analysis for MPG.

| **Models** | | **Training error** | **Test error** | |
|---|---|---|---|---|
| Linguistic model [28] | One-step optimization | 2.90 ± 0.52 | 3.17 ± 1.01 | RMSE |
| | Multi-step optimization | 2.86 ± 0.83 | 3.14 ± 1.01 | |
| RRbFM [29] | Growing 2 rules | 2.328 ± 0.095 | 3.010 ± 0.119 | RMSE |
| | Growing 3 rules | 2.605 ± 0.090 | 3.035 ± 0.136 | |
| Fuzzy model [30] | Random basis function | 3.047 ± 0.086 | 3.180 ± 0.455 | RMSE |
| DFCCNNs [31] | FCNNs with LSE | 2.785 ± 0.054 | 2.859 ± 0.494 | RMSE |
| | FCNNs with WLSE | 2.749 ± 0.073 | 2.863 ± 0.613 | |
| | DFCCNNs | 2.668 ± 0.055 | 2.748 ± 0.487 | |
| Proposed model | | 2.370± 0.047 | 2.702± 0.178 | RMSE |

### 5.2.3. Computer Activity data (CA)

We consider the CA dataset, which is a collection of computer system activity metrics. The dataset contains 21 input variables and 8192 pairs of input-output data patterns. The output is the fraction of time that CPUs run in user mode. Table 5 offers the performance of the control and the comparative models. MSE/2 is utilized as the performance index. The performance value is expressed in terms of mean and STD, and boldface indicates the best performance. Compared with zero-order FRM and first-order FRM, second-order FRM has better approximation ability but has worse test error (i.e., overfitting). A suitable penalty term can alleviate the overfitting of the model and improve the generalization ability of the model. In Table 5, the best prediction performance of RSFRM is 3.733 ± 0.248, which is significantly improved when compared with the three second-order FRMs with $\ell_2$.

Table 5 Experimental results of comparative and proposed models (CA)

| Models | No. of rules ($c$) | Penalty terms ($\lambda_1, \lambda_2, \lambda_3$) | Training error Mean | STD | Test error Mean | STD |
|---|---|---|---|---|---|---|
| Zero-order FRM | 10 | N/A | 124.9 | 4.121 | 125.0 | 16.86 |
| Linear-order FRM | 10 | N/A | 14.19 | 0.600 | 15.67 | 1.777 |
| Second-order FRM | 10 | N/A | 1.737 | 0.037 | 7115 | 12743 |
| Second-order FRM with $\ell_2(\lambda_1/\lambda_2/\lambda_3)$ | 10 | $10^{-8}$ | 1.729 | 0.041 | 13548 | 26638 |
| | 10 | $10^{-1}$ | 2.375 | 0.039 | 4.679 | 0.393 |
| | 10 | $10^1$ | 7.020 | 0.170 | 10.59 | 1.003 |
| RSFRM (Proposed model) | 10 | ($10^{-8}, 10^{-1}, 10^1$) | **3.011** | **0.045** | **3.733** | **0.248** |

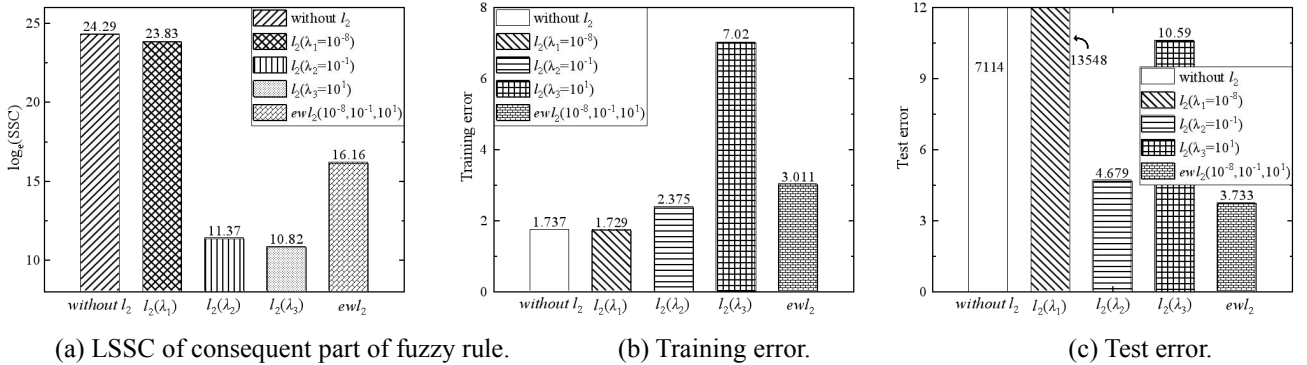

(a) LSSC of consequent part of fuzzy rule.  (b) Training error.  (c) Test error.

**Fig. 10.** Effects of regularization ($\ell_2$ and $ew$-$\ell_2$) on the coefficients and performance of models (CA). (LSSC: logarithm of the sum of the squared coefficients)

The effect of regularization techniques ($\ell_2$ and $ew$-$\ell_2$) on the coefficients and performance of the model is displayed in Fig. 10. In Fig. 10(a), the LSSC decreases as the penalty term ($\lambda$) increases (without $\ell_2$, $\lambda$ can be treated as zero). The growth of $\lambda$ does not always cause an increase in the training error of the model. When $\lambda = 10^{-8}$, the training error of the model is slightly lower than that without regularization, which makes the generalization performance worse (the test error is more lager). This is because the further training (overtraining) of the model makes its overfitting further serious. The proposed model with $ew$-$\ell_2$ can improve generalization ability by establishing a reasonable and effective trade-off between the training and testing of the model, and further exerting the representation ability of quadratic polynomials within local models.

Fig. 11 visualizes the impact of the complexity (represented by the number of fuzzy rules) on model performance. The black solid line with the circle represents the change of the proposed model (which uses $ew$-$\ell_2$) with the increasing of the complexity. When the penalty term ($\lambda$) is small (e.g., $10^{-8}, 10^{-1}$), the training error of the model decreases as the model complexity increases, and the larger penalty term (e.g., $10^1$) will increase the training error of the model. For small or large penalty terms (e.g., $10^{-8}$, $10^1$), the test error of the corresponding model becomes worse as the model complexity increases, and the appropriate λ can improve the generalization ability of the model. Compared with $\ell_2$, $ew$-$\ell_2$, which contains three different penalty terms, can better balance the relationship between the training performance and test performance of the model. As a result, $ew$-$\ell_2$ exhibits more excellent and stable prediction performance under different complexity. Table 6 offers a comparative summary of the proposed model compared to other models.

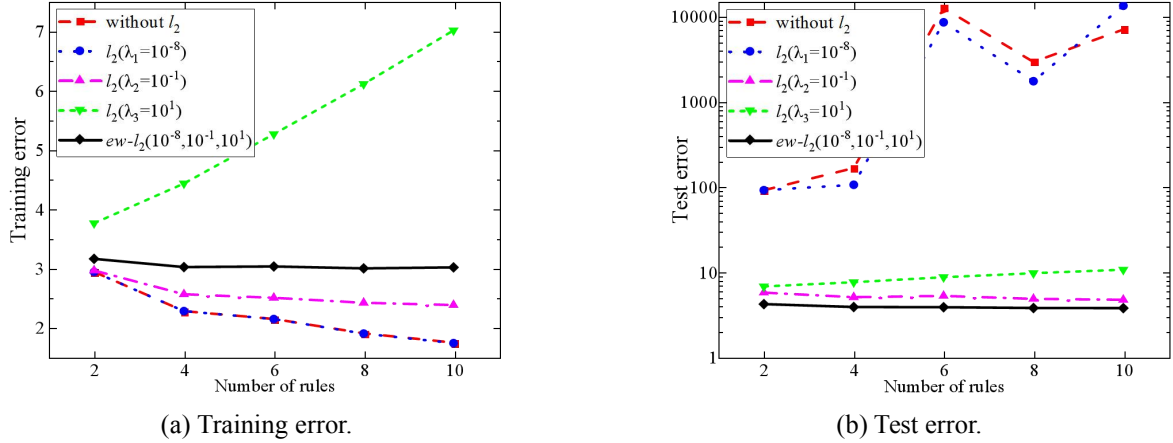

(a) Training error.  (b) Test error.

**Fig. 11.** Effects of complexity (number of rules) on the performance of models (CA).

Table 6 Results of comparative analysis for CA

| Models | | Training error | Test error | |
|---|---|---|---|---|
| MEA-FIS [32] | PAES$_{RB}$ | 18.21 | 19.27 | MSE/2 |
| | PAES$_{KB}$ | 11.99 | 13.43 | |
| MGA-FIS [33] | FS$_{MOGFS}$+T$_{UN}$ | 4.763±0.404 | 5.063±0.760 | MSE/2 |
| | FS$_{MOGFS}^e$+T$_{UN}^e$ | 5.021±0.422 | 5.216±0.483 | |
| METSK-HD$^e$ [34] | | 4.376 | 4.949 | MSE/2 |
| FRULER [35] | | | 4.634 | MSE/2 |
| MOKBL±MOMs [24] | Before MOMs | 7.43 | 7.91 | MSE/2 |
| | After MOMs | 4.52 | 4.67 | |
| Proposed model | | 3.011± 0.045 | 3.733± 0.248 | MSE/2 |

### 5.3. Overall evaluations of the proposed model

In order to further evaluate and analyze the prediction effect of the proposed model, the proposed model is experimented by using a series of machine learning datasets (http://keel.es/). Table 7 offers the pertinent details of the 23 datasets used in the experiments in this section, such as the abbreviation of the dataset, the number of dimensionalities, and the scale of the data. In order to make a fair comparison, we used the method of statistical analysis. The Friedman test is widely known as the nonparametric alternative of repeated measures analysis of variance, which is used to check repeated observations on the same subject and evaluate whether the measured average grade (rank) is remarkably different from the mean grade [36].

Table 8 shows the prediction performance (represented by test error) comparison between the comparative models including the second-order fuzzy rule-based model (FRM) and its three variants
16 / 22

with different $\ell_2$ penalty terms and the proposed model with $ew\text{-}\ell_2$. The numbers in parentheses indicate the rank of each model on the same dataset. When the significance level ($\alpha$) is set to 0.05, we can calculate the statistic $F_F$ according to the average rank (Avg. rank) shown in Table 8. Since $F_F$ = 18.50 is greater than the critical value F (4, 88) = 2.48, the Friedman test rejects the null hypothesis. For further comparison, we utilize the Bonferroni-Dunn test as the ad-hoc test to check whether the proposed model is statistically significantly superior to the other models. The Bonferroni-Dunn test is applicable to all comparative models are compared with the control (proposed) model without comparing between them. When the difference between the average ranks of two models (one is the control model and the other is the comparative model) is greater than at least the critical difference (CD), there is a significant difference in performance (that is, one model is significantly superior to the other) [33]. Under this experimental condition, the CD is 1.16. According to the results obtained in Table 9, the differences between the control model and the four comparison models are all greater than (or equal to) CD. We can conclude that the exponential weighted $\ell_2$ regularization ($ew\text{-}\ell_2$) can significantly improve the generalization ability of the model compared with ordinary $\ell_2$ and without $\ell_2$.

Table 7 Summary of datasets with various complexity

| NO. | Dataset | Abbreviation | No. of dimensions | Scale of data |
|---|---|---|---|---|
| 1 | Electrical length | EL1 | 2 | 495 |
| 2 | Plastic | PLC | 2 | 1650 |
| 3 | Quake | QU | 3 | 2178 |
| 4 | Electrical Maintenance | EL2 | 4 | 1056 |
| 5 | Friedman Benchmark Function | FRI | 5 | 1201 |
| 6 | AutoMPG6 | MP6 | 5 | 392 |
| 7 | Delta Ailerons | DA | 5 | 7129 |
| 8 | Daily Electricity Energy | DEE | 6 | 365 |
| 9 | Delta Elevators | DE | 6 | 9517 |
| 10 | Analizing Categorical | AC | 7 | 4052 |
| 11 | Automobile Miles Per Gallon | MPG | 7 | 392 |
| 12 | Abalone | ABA | 8 | 4177 |
| 13 | Concrete compressive strength | CCS | 8 | 1030 |
| 14 | Stock Prices | STP | 9 | 950 |
| 15 | Weather Ankara | WAN | 9 | 1609 |
| 16 | Weather Izmir | WIZ | 9 | 1461 |
| 17 | California Housing | CAL | 8 | 20640 |
| 18 | Forest Fires | FF | 12 | 517 |
| 19 | Mortgage | MOR | 15 | 1049 |
| 20 | Treasury | TRE | 15 | 1049 |
| 21 | Baseball | BAS | 16 | 337 |
| 22 | Computer Activity | CA | 21 | 8192 |
| 23 | Elevators | ELV | 18 | 16599 |

In addition, this study applied three state-of-art regression models such as MOKBL±MOMs [24],

FRULER [35] and METSK-HD[e] [34] to compare the overall performance of the proposed model, including prediction performance and complexity. We use the number of fuzzy rules in the model to quantify the complexity of the model. It is generally believed that the fewer the number of rules, the stronger the interpretability of the model [23]. Table 10 shows the prediction ability (which is denoted by test error) and the number of rules of the models constructed on the various datasets of the proposed and the comparative models. MSE/2 is utilized as the performance index. The entries in boldface mean the smallest test error (or the number of rules) among the entire models. We use the Friedman test to make a fair comparison of the model's prediction performance and complexity, respectively. The last row of the table represents the average rank of the test errors and number of rules of each model on all datasets.

Table 8 Performance comparison between proposed model and comparative models. (Errors in this table should be multiplied by $10^{-5}$, $10^{-8}$, $10^{-6}$, $10^{5}$, $10^{9}$, $10^{-6}$ in the case of EL1, DA, DE, BAS, CAL, ELV, respectively.)

| Data | Second-order FRM without/ with $\ell_2$ | | | | Proposed model |
|---|---|---|---|---|---|
| | without $\ell_2$ | with $\ell_2(\lambda_1)$ | with $\ell_2(\lambda_2)$ | with $\ell_2(\lambda_3)$ | with $ew$-$\ell_2$ |
| EL1 | 2.067 (4) | 2.040 (2.5) | 2.040 (2.5) | 6.691 (5) | **1.874 (1)** |
| PLC | **1.130 (1)** | 1.133 (4.5) | 1.133 (4.5) | 1.132 (2.5) | 1.132 (2.5) |
| QU | **0.0179 (1.5)** | 0.0191 (3) | 0.3382 (4) | 2.4153 (5) | **0.0179 (1.5)** |
| EL2 | 158274 (5) | 54555 (4) | 4972 (3) | 4666 (2) | **4353 (1)** |
| FRI | 17.03 (5) | **0.803 (1)** | 0.898 (2) | 1.410 (4) | 1.350 (3) |
| MP6 | 13.32 (5) | 4.335 (4) | 3.926 (2) | 4.307 (3) | **3.692 (1)** |
| DA | **1.325 (2.5)** | **1.325 (2.5)** | **1.325 (2.5)** | 1.326 (5) | **1.325 (2.5)** |
| DEE | 0.454 (5) | 0.397 (4) | 0.108 (2) | 0.248 (3) | **0.079 (1)** |
| DE | 868.1 (5) | **1.006 (2)** | **1.006 (2)** | 1.008 (4) | **1.006 (2)** |
| AC | 21505 (5) | 0.132 (4) | 0.012 (3) | **0.008 (1)** | 0.009 (2) |
| MPG | 38891 (5) | 4.143 (2) | 4.395 (3) | 9.636 (4) | **3.662 (1)** |
| ABA | 7.691 (4) | 3.785 (3) | 3.384 (2) | 13.40 (5) | **2.277 (1)** |
| CCS | 46345 (5) | **26.72 (1)** | 26.74 (2) | 27.81 (4) | 27.54 (3) |
| STP | 8.607 (5) | 0.526 (4) | 0.329 (2) | 0.518 (3) | **0.275 (1)** |
| WAN | 1.449 (5) | 0.813 (2.5) | 0.813 (2.5) | 0.915 (4) | **0.734 (1)** |
| WIZ | 0.896 (4) | 0.807 (2.5) | 0.807 (2.5) | 6.192 (5) | **0.662 (1)** |
| CAL | 2.370 (4) | 2.052 (2) | 2.138 (3) | 4.214 (5) | **2.039 (1)** |
| FF | 749436 (5) | 14594 (4) | 4121 (3) | **2094 (1)** | 2137 (2) |
| MOR | 27.901 (5) | 0.1942 (4) | 0.0063 (3) | 0.0037 (2) | **0.0027 (1)** |
| TRE | 151.28 (5) | 0.4980 (4) | 0.0187 (2) | 0.0202 (3) | **0.0165 (1)** |
| BAS | 1658 (5) | 910.1 (4) | 5.108 (3) | 5.008 (2) | **2.215 (1)** |
| CA | 7114 (4) | 13548 (5) | 4.679 (2) | 10.59 (3) | **3.733 (1)** |
| ELV | 2014 (5) | 27.94 (4) | 21.68 (3) | 12.74 (2) | **2.970 (1)** |
| Avg. rank | 4.35 | 3.20 | 2.63 | 3.37 | 1.46 |

(A) Comparison of prediction ability. Because of $F_F$ = 4.03 > F (3,66) = 2.74, the null hypothesis

is rejected. We use Bonferroni-Dunn test for further analysis and comparison. If the difference between the average ranks of the two models is at least greater than one CD, then there is a clear difference between the two models. Under this experimental condition, CD is 0.91. In Table 11, the difference in the average rank between all the comparative models and the control model is greater than CD. It can be argued that the proposed RSFRM is superior to the three state-of-art models in terms of prediction ability.

Table 9 Difference of the Avg. rank of between proposed models and comparative models

| Difference | Second-order FRM without/ with $\ell_2$ | | | |
|---|---|---|---|---|
| | without $\ell_2$ | with $\ell_2(\lambda_1)$ | with $\ell_2(\lambda_2)$ | with $\ell_2(\lambda_3)$ |
| **Proposed model (with $ew\text{-}\ell_2$)** | 2.89 (> CD) | 1.74 (> CD) | 1.17 (> CD) | 1.91 (> CD) |

Table 10 Performance comparison between the proposed RSFRM and state-of-art models. (Errors in the table should be multiplied by $10^{-5}$, $10^{-8}$, $10^{-6}$, $10^5$, $10^9$, $10^{-6}$ in the case of EL1, DA, DE, BAS, CAL, ELV, respectively.)

| Data | MOKBL±MOMs [21] | | FRULER [32] | | METSK-HD$^e$ [31] | | Proposed RSFRM | |
|---|---|---|---|---|---|---|---|---|
| | Rules | Test error | Rules | Test error | Rules | Test error | Rules | Test error |
| **EL1** | 4.8 | **1.87** | 4.1 | 2.012 | 11.4 | 2.022 | **4** | 1.874 |
| **PLC** | 7.1 | 1.181 | **1.4** | 1.219 | 19.2 | 1.136 | 6 | **1.132** |
| **QU** | 7.8 | **0.0170** | 7.8 | 0.0181 | 18.3 | 0.0181 | **2** | 0.0179 |
| **EL2** | 10.9 | 12733 | **4.3** | 6729 | 36.9 | **3192** | 10 | 4353 |
| **FRI** | 13 | 2.74 | 8.0 | **0.731** | 66 | 1.888 | **4** | 1.449 |
| **MP6** | 10 | 4.51 | 13.7 | 3.727 | 53.6 | 4.478 | **8** | **3.692** |
| **DA** | 9.5 | 1.92 | **2.5** | 1.458 | 36.8 | 1.402 | 4 | **1.325** |
| **DEE** | 8.3 | 0.088 | **7.9** | 0.080 | 50.6 | 0.103 | 8 | **0.079** |
| **DE** | 6.1 | 1.407 | **5.8** | 1.045 | 39.1 | 1.031 | 6 | **1.006** |
| **AC** | 9.3 | 0.008 | **3.9** | 0.008 | 33.3 | **0.004** | 10 | 0.009 |
| **MPG** | 12.1 | 4.24 | 12.7 | 4.084 | 64.2 | 5.391 | **8** | **3.662** |
| **ABA** | 6.8 | 2.401 | **4.5** | 2.393 | 23.1 | 2.392 | 8 | **2.277** |
| **CCS** | 10.2 | 27.42 | 8.9 | **20.598** | 53.7 | 23.885 | **4** | 27.223 |
| **STP** | 11.9 | 0.66 | 42.4 | 0.353 | 66.4 | 0.387 | **10** | **0.275** |
| **WAN** | 7.2 | 1.60 | 5.6 | 0.888 | 48 | 1.189 | **2** | **0.734** |
| **WIZ** | 7.8 | 1.58 | 8.9 | 0.663 | 29.1 | 0.944 | **2** | **0.662** |
| **CAL** | **7.4** | 2.66 | 15.4 | 2.110 | 55.8 | **1.71** | 8 | 2.039 |
| **FF** | **3.9** | **2006** | 5.6 | 2214 | 40.6 | 5587 | 4 | 2137 |
| **MOR** | 9.5 | 0.015 | **7.9** | 0.007 | 27.2 | 0.013 | 8 | **0.0027** |
| **TRE** | 5.3 | 0.041 | **4.5** | 0.027 | 28.1 | 0.038 | 6 | **0.0165** |
| **BAS** | 9.3 | 2.57 | 6.2 | 3.0578 | 59.8 | 3.688 | **2** | **2.215** |
| **CA** | 15.5 | 4.67 | **7.1** | 4.634 | 32.91 | 4.949 | 10 | **3.733** |
| **ELV** | 14 | 10.7 | 5.4 | **2.93** | 34.9 | 7.02 | **4** | 2.97 |
| **Avg. rank** | 2.59 | 3.11 | 1.76 | 2.44 | 4.00 | 2.80 | 1.65 | 1.52 |

Table 11 Difference of the Avg. rank of between proposed models and comparative models

| Difference | MOKBL±MOMs [21] | FRULER [32] | METSK-HD$^e$ [31] |
|---|---|---|---|
| **Proposed RSFRM (for test error)** | 1.59 (> CD) | 0.92 (> CD) | 1.28 (> CD) |
| **Proposed RSFRM (for rules)** | 0.94 (> CD) | 0.11 (< CD) | 2.35 (> CD) |

(B) Comparison of complexity. Under the condition that $F_F = 52.74 > F(3,66) = 2.74$, the Friedman test rejects the null hypothesis. The third row of Table 11 shows the difference in the average rank of the number of rules between the proposed RSFRM and the three comparative models. From the Bonferroni-Dunn test (CD is 0.91), we can say that the proposed RSFRM is lower to METSK-HD$^e$ and MOKBL±MOMs in terms of complexity (low complexity means high interpretability). Through the experimental results reported in Table 10, the complexity of the proposed RSFRM is the similar as that of FRULER (win: 12 datasets, loss: 11 datasets).

## 6. Concluding remarks

In this article, we have presented and investigated the reinforced second-order fuzzy rule-based model (RSFRM), which is realized with the aid of Fuzzy C-Means (FCM) partition and quadratic polynomial (QP) as well as exponential weighted $\ell_2$ regularization ($ew\text{-}\ell_2$). FCM is utilized to analyze data distribution and divide the input space into local regions, as well as form the antecedent parts of fuzzy rules. The consequent part of the fuzzy rule is constructed by the quadratic polynomial. Compared with constant and linear function, QP can better describe the behavior within the local region by refining the relationship between input and output. $ew\text{-}\ell_2$ is designed to alleviate the overfitting problem that QP may cause. Different from the ordinary $\ell_2$, $ew\text{-}\ell_2$ can separately identify and penalize the terms of different types of polynomial terms in the coefficient estimation, and its results can not only alleviate the overfitting and prevent the deterioration of generalization ability but also effectively release the prediction potential of the model.

The proposed RSFRM was comprehensively evaluated and statistically analyzed by using 23 machine learning datasets with different complexity. Through the experimental results, we proved that the prediction ability of RSFRM is better than that of its related comparative models. We also compared RSFRM with three start-of-art models, and the results show that RSFRM boosts the prediction accuracy of the model without sacrificing the interpretability of the model (that is, without increasing the complexity of the model).

The proposed RSFRM does not consider the influence of feature selection and dimensionality reduction on the performance of the model. Therefore, future work includes selecting appropriate feature selection or dimensionality reduction techniques to further strengthen the performance of the model.